# On Efficient Real-Time Semantic Segmentation: A Survey

Christopher J. Holder, Muhammad Shafique

**Abstract**—Semantic segmentation is the problem of assigning a class label to every pixel in an image, and is an important component of the vision stack of autonomous mobile systems for facilitating scene understanding and object detection. However, many of the top performing semantic segmentation models are extremely complex and cumbersome, and as such are not suited to deployment onboard mobile platforms where computational resources are limited and low-latency operation is a vital requirement. In this survey, we take a thorough look at the works that aim to address this misalignment with more compact and efficient models capable of deployment on low-memory embedded systems while meeting the constraint of real-time inference. We discuss several of the most prominent works in the field, placing them within a taxonomy based on their major contributions, and finally we evaluate the inference speed of the discussed models under consistent hardware and software setups that represent a typical research environment with high-end GPU and a realistic deployed scenario using low-memory embedded GPU hardware. Our experimental results demonstrate that many works are capable of real-time performance on resource-constrained hardware, while illustrating the consistent trade-off between latency and accuracy.

—————————— ◆ ——————————

## 1 INTRODUCTION

SEMANTIC segmentation is the act of dividing spatially structured data, such as 2D images, volumetric 3D representations or spatiotemporal video sequences, into semantically meaningful regions, and has applications in a wide variety of fields, including medical diagnosis [1], human-computer interaction [2], and robotics. As an important step towards scene understanding for autonomous agents such as robots and self-driving cars, many real-world applications of semantic segmentation are required to operate in real-time on heavily memory and resource-constrained hardware, however much of the current state-of-the-art neglects this in the pursuit of ever-greater performance on high-powered multi-GPU systems. This survey aims to address this by exploring any prior works whose aims include low-latency or low-memory inference, even if such constraints lead to a compromise in overall accuracy. As such, we believe this is the first work to comprehensively review the field of low-latency deep learning architectures for semantic segmentation, and more notably the first to include an experimental comparison of notable works under consistent hardware and software conditions. In this survey, we specifically focus on works that aim to segment individual 2D colour images – i.e. not 3D volumetric data, video sequences, grayscale or multi-spectral inputs – into multiple semantic classes, with a latency low enough to facilitate real-time inference, which we define as >= 30fps. Not only do we discuss the major contributions of these works, but we perform our own inference latency evaluation of 24 notable works published between 2015 and 2021 under consistent hardware and software conditions, addressing the problem that model runtimes quoted in the original papers have been recorded on wildly different setups and thus cannot be objectively compared. We evaluate works under both high-end research workstation, and constrained embedded device scenarios to better assess suitability to real-world deployment.
The paper is organized as shown in Fig. 1, with the content of subsequent sections detailed as follows:

• C.J. Holder and M. Shafique are with the Division of Engineering, New York University Abu Dhabi, P.O. Box 129188, Saadiyat Island, Abu Dhabi, United Arab Emirates. E-mail: {chris.holder, ms12713}@nyu.edu.

**§1. Introduction**
**§2. Background**
  –§2.1. Semantic Segmentation
  –§2.2. Challenges and Trade-offs in Designing Deep Semantic Segmentation Architectures
  –§2.3. Applications
  –§2.4. Datasets
**§3. General Techniques for Improving Efficiency of Deep CNNs**
  –§3.1. Downsampling and Upsampling
  –§3.2. Efficient Convolution
  –§3.3. Residual Connections
  –§3.4. Backbone Architectures
**§4. Taxonomy of Prominent Works**
  –§4.1. Encoder-Decoder
  –§4.2. Multi-Branch
  –§4.3. Meta-Learning
  –§4.4. Attention
  –§4.5. Training Pipeline
**§5. Evaluation**
**§6. Conclusions**

Fig. 1. Organisation of the paper

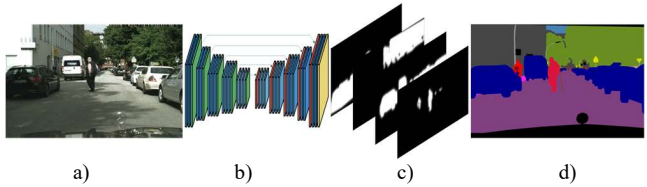

<div>a)      b)      c)      d)</div>

Fig. 2. An overview of the semantic segmentation pipeline: A 2D colour image (a) is input to a CNN (b), in this case SegNet [18], whose output is $C$ class probability maps (c), where $C$ is the number of semantic classes to be predicted, from which the final segmentation result (d) can be derived via an argmax operation.

- The **Background** section discusses the development of the semantic segmentation problem, including significant works, applications, datasets and challenges.

- The **Techniques** section lists several of the key methods utilized to achieve real-time semantic segmentation performance.

- The **Taxonomy of Prominent Works** divides approaches into five categories, and discusses the contributions of each in depth.

- The **Evaluation** section compares the major works in the field by their performance on commonly used datasets, as well as inference latency observed in our own experiments on both high-end and embedded hardware.

- The **Conclusions** draw the survey to a close, discussing overarching themes and challenges that remain to be addressed in the field of real-time semantic segmentation.

## 2 BACKGROUND

In this section, we briefly cover the history of semantic segmentation, including methods, datasets and applications and challenges, and discuss the importance, and difficulty, of achieving accurate results in real-time on constrained computation platforms.

### 2.1 Semantic Segmentation

Early image segmentation approaches were capable of dividing images into regions based on little more than basic colour and low-level textural information [3] [4]. Such techniques could be combined with machine learning methods such as Support Vector Machine [5] or Random Forest [6], commonly in a process involving the segmentation of an image into superpixels that are subsequently classified [7] [8], however performance was limited by the accuracy of a segmentation method naïve to semantic information as well as the limitations of the machine learning algorithm used. With the popularization of deep-learning techniques in the 2010s, due in part to the availability of massively parallel GPUs and large labelled datasets, came the potential for convolutional neural networks (CNNs) capable of combining colour, texture, and semantic information to produce significantly more accurate results. Early deep-learning-based vision models were often focused on classification [9], wherein a single image is assigned a single semantic label. The more fine-grained task of detection, wherein multiple objects are classified and located within an image subsequently garnered a large amount of research interest [10] [11]. This quest for ever finer-grained scene understanding paradigms eventually led to the emergence of semantic segmentation, illustrated in Fig. 2. A CNN assigns every pixel in an input image an output vector of class probabilities, wherein each element denotes the likelihood of the given pixel belonging to one of a predetermined number of semantic classes, with the class assigned the maximum likelihood taken to be the predicted class label for that pixel. Subsequent works have further built on these ideas, including instance segmentation [12], wherein pixels are labelled not just by semantic class but by individual instance of that class, volumetric segmentation, wherein each component of a 3D scene, which may be represented as a point cloud [13], mesh [14], or voxel grid [15], is assigned a class label, and video segmentation [16], wherein semantic classes are tracked through a temporal sequence of images, however these are beyond the scope of this survey.

### 2.2 Challenges and Trade-offs in Designing Deep Semantic Segmentation Architectures

Deep-learning architectures designed for classification and detection tasks typically extract features through a sequence of convolutions followed by one or more fully-connected operators that aggregate feature maps into an output vector, discarding structural information in the process. As semantic segmentation relies on structural information and aims to produce an output of the same dimensions as the input image, early works followed the paradigm of the Fully Convolutional Network (FCN) [17], in which the fully connected operations are discarded. While FCN architectures are capable of retaining the structural information needed to produce accurate segmentation results, the process of extracting meaningful semantic information typically involves passing an input image, and its derived feature maps, through multiple successive downsampling operations, in turn discarding much of the fine-grained detail required to accurately infer class boundaries in the subsequent upsampled output. This challenge of extracting high-level global context while retaining low-level details has formed a major theme of the subsequent development of the semantic segmentation field. Attempts to address this have included storing and reusing the indices used by pooling operations to downsample [18], reusing early high-resolution feature maps to refine the output [1], and multi-branch architectures [19] that fuse high- and low- level features to produce the final output, however the search for ever more accurate results often leads to highly complex architectures that are computationally expensive and slow.

### 2.3 Applications

Many real-world applications of semantic segmentation models require that they be run with very little latency on mobile devices that have limited memory and computational power. The scene understanding capabilities that such models facilitate are of particular use to autonomous platforms such as robots [20] [21], drones [22] [23], and self-driving cars [18] [24], where cumbersome architectures that may have an inference latency measured in seconds are not viable. For this reason, there is significant research and industrial interest in models capable of accurate semantic segmentation with low memory requirements and high inference speed, particularly those capable of running on Nvidia's Tegra embedded GPUs [25], whose 'Jetson' variant is used with many robotic and drone platforms, and whose Drive variant is designed for the automotive industry. Other applications of semantic segmentation include human-computer interaction [2], where the low-latency constraint is still important, and medical diagnosis [1], where latency is less important but where memory may still be constrained if a model is to be run directly on a diagnostic imaging device, for example.

TABLE 1
DETAILS OF SEVERAL SEMANTIC SEGMENTATION DATASETS

| Dataset | Scenario | Published | Labelled images | Resolution |
|---|---|---|---|---|
| CamVid [26] | Autonomous Driving | 2009 | 701 | 960×720 |
| KITTI [27] | Autonomous Driving | 2012 | 400 | 1242×375 |
| Cityscapes [28] | Autonomous Driving | 2016 | 5000/20,000 (coarse) | 2048×1024 |
| Berkeley DeepDrive [29] | Autonomous Driving | 2018 | 10,000 | 1280×720 |
| Audi Autonomous Driving [30] | Autonomous Driving | 2020 | 41,280 | 1920×1208 |
| PASCAL VOC2012 [31] | Objects | 2012 | 9993 | variable |
| NYU Depth V2 [32] | Indoor Scenes | 2012 | 1449 | 640×480 |

## 2.4 Datasets

Further emphasizing the applicability of semantic segmentation to autonomous vehicles, several prominent datasets used to benchmark proposed architectures comprise images of driving scenarios, some examples of which are shown in Fig. 3, with details listed in Table I. In 2009, **CamVid** [26] presented the first densely labelled driving dataset, comprising 701 images from the perspective of a car driver, taken from a 10-minute sequence of driving in Cambridge, UK, and labelled with 32 semantic classes relevant to driving, including traffic light, car, and pedestrian. The relatively small number of images and low resolution of 960×720 makes CamVid a useful tool for rapid prototyping as a deep CNN can be trained to convergence in a relatively short period of time, and as such CamVid results are still widely quoted in contemporary works. The **KITTI** dataset [27], originally released in 2012 but updated many times since, includes dense semantic labels for just 400 images of German road scenes, however these are combined with stereoscopic, LIDAR, odometry, object detection and tracking, and instance segmentation labels, and so the dataset is widely used to benchmark a variety of driving related challenges. The **Cityscapes** dataset [28], published in 2016, contains 5000 densely labelled, and 20,000 coarsely labelled, images covering both semantic and instance segmentation at a resolution of 2048×1024, captured in several German cities, and has become the de-facto benchmark for automotive scene understanding. **Berkeley DeepDrive** [29], released in 2018, is made up of 100,000 videos of US roads covering 10 different tasks, including lane detection and object tracking, and includes dense semantic labels covering 40 classes across 10,000 images. The **Audi**

**Autonomous Driving Dataset (A2D2)** [30], published in 2020, includes 41,280 densely labelled images of German roads covering 38 semantic classes, with LIDAR used to also generate 3D bounding boxes and pointclouds for 3D segmentation.

Other semantic segmentation datasets include PASCAL VOC2012 [31], which includes labels for 20 object classes across 9993 images, although unlike other datasets that consider full scene understanding, image regions that are not a foreground object all fall into a single 'background' class. **NYU Depth V2** [32] comprises 1449 densely labelled RGB-D depth images captured by Microsoft Kinect [33] in a variety of indoor scenes. For evaluating real-time semantic segmentation approaches in this survey, we focus on results from CamVid and Cityscapes, as these are the most often quoted in such works.

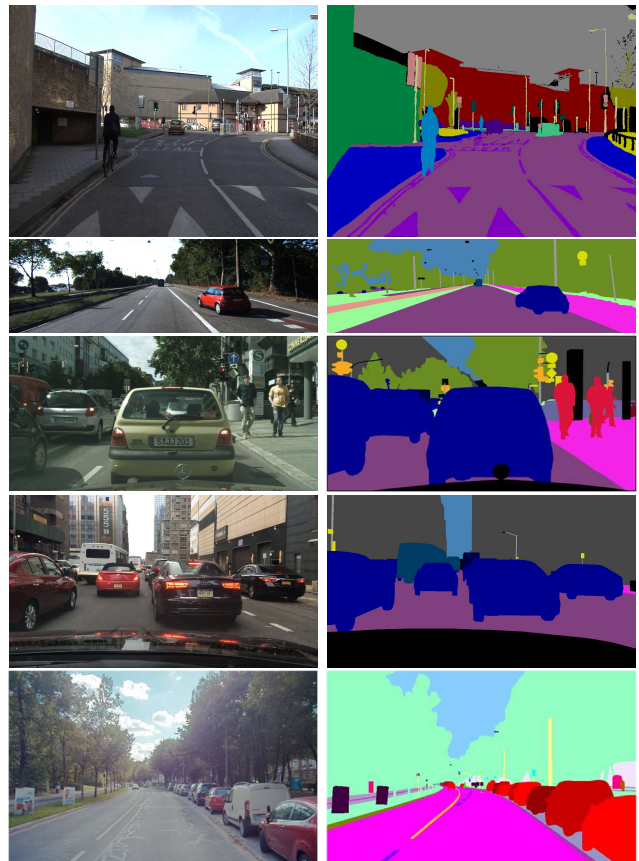

Fig. 3. Example images and ground truth segmentation from the (top to bottom) CamVid [26], KITTI [27], Cityscapes [28], Berkeley Deep-Drive [29], and A2D2 [30] datasets. The A2D2 example includes instance as well as semantic labels.

TABLE 2
EFFICIENT CONVOLUTIONAL OPERATORS

| Convolution Type | Parameters | Receptive Window Size |
|---|---|---|
| Standard | $k^2 \times C_1 \times C_2$ | $k$ |
| Depthwise Separable | $C_1 \times C_2 + k^2 \times C_2$ | $k$ |
| Grouped | $k^2 \times C_1 \times C_2 / g$ | $k$ |
| Asymmetric | $2 \times k \times C_1 \times C_2$ | $k$ |
| Bottleneck | $C_1 \times C_6 + k^2 \times C_2^2 + C_6 \times C_2$ | $k$ |
| Dilated | $k^2 \times C_1 \times C_2$ | $k \times d$-1 |

## 3 GENERAL TECHNIQUES FOR IMPROVING EFFICIENCY OF DEEP CNNS

In this section we list some of the commonly used techniques for decreasing the latency and parameter count of semantic segmentation architectures, and of deep neural networks in general, many of which have been exploited by several of the works discussed in the next section.

### 3.1 Downsampling and Upsampling

The number of operations required to apply a convolution to an image or feature map is proportional to its resolution, and so the inference latency of a CNN can be significantly reduced by downsampling input images, usually at the expense of output accuracy. However, downsampling is also widely used in large, complex models to increase the receptive field of deeper kernels such that they can better extract global context information. Careful consideration of the placement and type of downsampling operations used in a network, as well as how prior extracted fine-grained detail is utilized, can facilitate more efficient architectures with minimal impact on performance. Typically, downsampling early leads to a more compact model, as in [34], while later downsampling enables better extraction of high-resolution detail, as in [35]. Commonly used operations for downsampling are max- or average-pooling, wherein the largest value or mean, respectively, from a predetermined kernel window are passed to the output, and strided convolution, which learns a downsampling function over a fixed kernel size.

In segmentation, unlike many other vision problems, it is generally expected that the output will match the dimensions of the input, so downsampled feature maps must be subsequently upsampled. To preserve efficiency, many of the discussed works feature extremely lightweight decoders [36] that use interpolation to upsample and minimal convolutions to compute the final class probability map or integrate feature maps at different scales. The simplest interpolation method is Nearest Neighbour, in which a single pixel value is propagated across a window, although this can lead to overly coarse boundaries. Bilinear interpolation is computationally more expensive, but creates a smoother output, although it still lacks any capability to restore lost fine-grained information. Fractionally-strided convolutions, also known as deconvolutions or transposed convolutions, upsample their input by applying a learned kernel across an image with a stride of < 1 such that the output is larger than the input, however this introduces extra parameters over interpolation, which may not be desirable in a compact architecture, and can cause grid artefacts due to the manner in which input pixels are implicitly zero-padded.

### 3.2 Efficient Convolution

In a typical convolution layer with a stride of 1, no bias, and padded such that input and output dimensions are the same, a kernel of size $k \times k$ is applied across an input of $C_1 \times H \times W$ to produce an output of $C_2 \times H \times W$, where $H$ and $W$ are height and width, and $C_1$ and $C_2$ are the number of channels in the input and output respectively. Such a layer is parameterized by $k^2 C_1 C_2$ weights, and requires $k^2 C_1 C_2 HW$ operations to compute. There are several techniques commonly used to reduce the number of parameters and computations required by the convolution layers of efficient CNNs, illustrated in Fig. 4 and detailed in Table II.

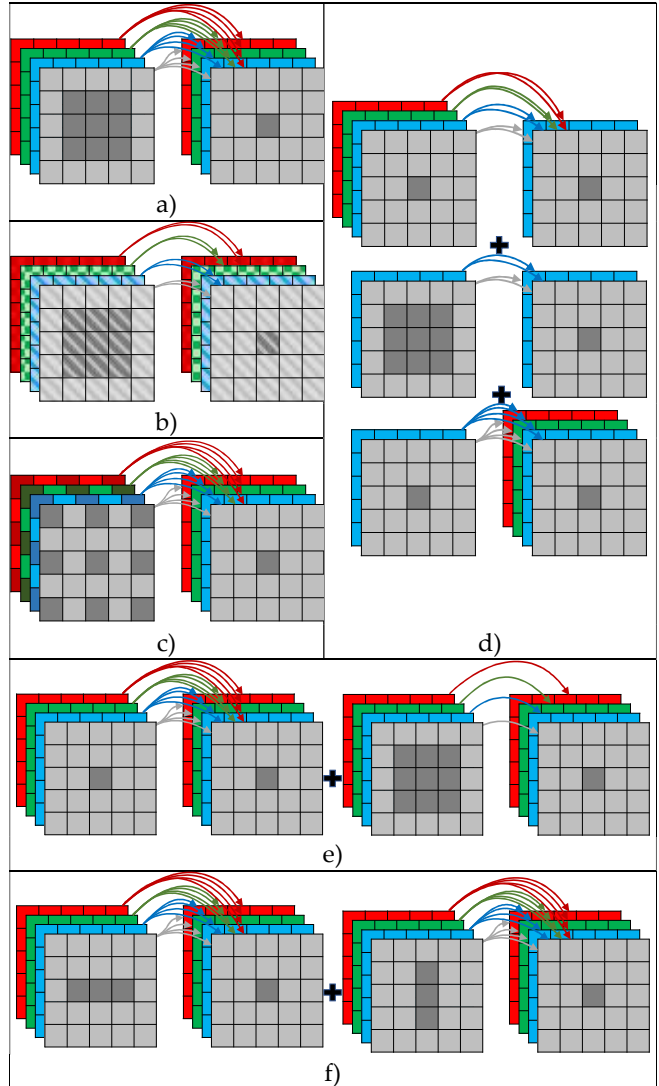

Fig. 4. Basic implementations of efficient convolution operators employed by many real-time semantic segmentation works: a) Standard 3×3 convolution, in which all output channels receive data from all input channels. b) Grouped convolution, in which output channels only receive from input channels within the same group (denoted by pattern). c) Dilated convolution whereby receptive field in increased without increasing parameters via a sparsely applied kernel. d) Bottleneck block, in which a 1×1 operator reduces the number of channels input to a standard 3×3 convolution before a 2nd 1×1 convolution expands the output. e) Depthwise-separable convolution, comprising a standard 1×1 operator and a 3×3 operator in which output channels only connect to a single input channel. f) Asymmetric Convolution comprising a 3×1 and a 1×3 operator. Shaded pixels in the input contribute to the shaded pixel in the output.

**Depthwise-Separable Convolution** [37] splits a convolution into a depth-wise and a point-wise convolution. The depth-wise convolution applies a single $k \times k$ kernel to each input channel to compute a single output channel, such that there is no communication between channels, requiring $k^2 C_1$ parameters and $k^2 C_1 HW$ operations. The point-wise convolution applies a $1 \times 1$ convolution linking all input channels to all output channels, requiring $C_1 C_2$ parameters and $C_1 C_2 HW$ operations. The two convolutions can be in either order [38], or a single depthwise may be placed between two pointwise convolutions [39].

**Grouped Convolution** [9] splits the channels of input and output into $g$ groups, with output filters only applied to input channels belonging to the corresponding group, reducing parameters and operations by a factor of $g$. One downside to grouped convolution is the lack of information sharing between groups. This was addressed in Shufflenet [40] with the Channel Shuffle operation in which groups of channels are further divided into sub-groups and reshaped such that channels are grouped differently for each convolution unit.

**Asymmetric Convolution** [41], or factorized convolution, refactors a $k \times k$ convolution as a $k \times 1$ and a $1 \times k$ convolution, requiring $2k C_1 C_2$ parameters and $2k C_1 C_2 HW$ operations.

**Bottleneck,** originally proposed as part of ResNet [42] involves using a $1 \times 1$ convolution to reduce the number of feature map channels so that subsequent larger-kernel convolutions are more efficient. Another 1x1 kernel reprojects the output back to the input dimensionality. Squeezenet [43] built on this with the 'Fire' module that combines multiple convolutions that 'squeeze' and then expand feature dimensionality.

**Dilated Convolution** [44], or atrous convolution, enable a larger receptive field without increasing kernel size by applying kernel weights sparsely over a larger input window. An additional parameter, dilation rate $d$, determines the size of input window over which a kernel is applied.

### 3.3 Residual Connections
Residual, or skip, connections allow data within a network to bypass certain operations. Such connections serve several purposes, including improving the flow of gradients during backpropagation and reuse of features from previous layers [42], and are commonly used in segmentation networks to refine low-resolution feature maps via fusion with high-resolution features from earlier layers [1]. Many contemporary CNN architectures employ a structure comprising a sequence of some variant of residual block, in which the output of the operations within the block is summed with the input.

### 3.4 Backbone Architectures
Many semantic segmentation models employ one of several widely used backbone networks as a feature extractor, often an architecture designed for, and sometimes pretrained on, a classification task, adapted to output features that can be upscaled for segmentation.

**ResNet** [42] demonstrated that residual connections can facilitate relatively easy optimization of very deep networks, and achieved state-of-the-art image classification results with fewer parameters than comparable models. Two of the basic building blocks of a ResNet architecture are depicted in Fig. 5a. **Squeezenet** [43], which was specifically designed to be compact and fast enough for real-time embedded systems, demonstrated that classification results comparable to prior works could be obtained from significantly smaller models via the squeezing and expansion of feature maps. **Shufflenet** [40] demonstrated group convolutions and channel shuffle could be used to gain respectable classification results from a network small enough to run on embedded systems. Shufflenet units are depicted in Fig. 5b. The original **Mobilenet** [45] made extensive use of depthwise-separable convolutions to demonstrate efficient classification performance. **MobileNetV2** [46] introduced the Inverted Residual Block, whereby feature dimensionality is increased before the depth-wise convolution, and subsequently reduced again, inverting the more common practice of reducing then expanding features. MobilenetV2 blocks are depicted in Fig 5c. **EfficientNet** [47] presented a family of compact architectures based on the MobileNetV2 Inverted Bottleneck wherein network depth, feature channels and resolution have been optimized by grid search for predetermined sizes of model.

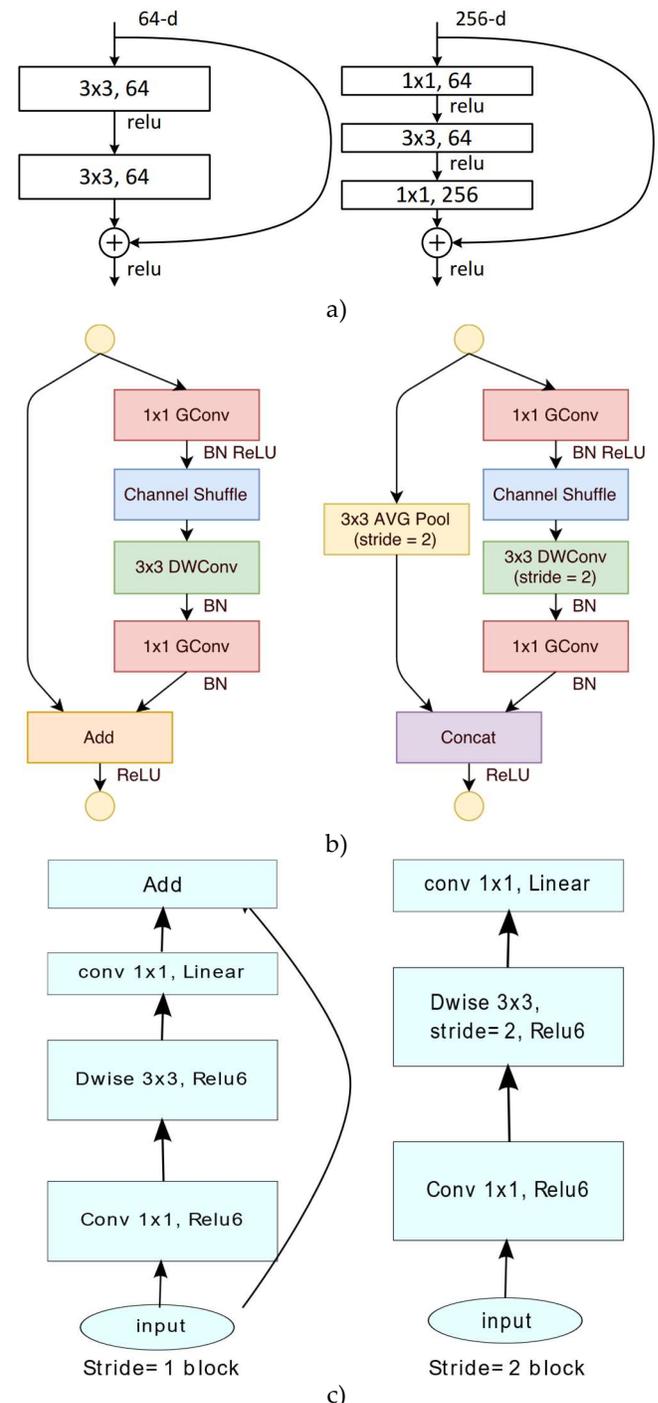

Fig. 5. Examples of convolution blocks from three common backbone architectures. a) ResNet [42] residual block and bottleneck; b) Shuflenet [40] unit with grouped convolution and channel shuffle, and strided Shufflenet unit that downsamples input by a factor of 2; c) MobilenetV2 [46] block using depthwise separable convolution, and downsampling block where the 3x3 convolution uses a stride of 2 and no residual connection.

TABLE 3
Taxonomy Categories

| Category | Description | Works |
|---|---|---|
| Encoder-Decoder | 2-stage network that first encodes input features, and subsequently decodes them to generate output | [1] [18] [34] [35] [51] [52] [53] [55] [57] [59] [61] [62] [63] [64] [65] [66] [69] |
| Multi-Branch | A model that processes inputs at 2 or more resolutions | [70] [39] [72] [19] [74] |
| Meta-Learning | Model architecture/weights are set by a separate learned model | [79] [83] [84] [87] |
| Attention | Global context is aggregated by an attention-based method | [36] [92] [93] [94] |
| Training Pipeline | Existing architectures improved by better training/optimisation process | [96] [99] [100] |

## 4 Taxonomy of Prominent Works

In this section we discuss many of the varied works proposed to address the challenge of real time semantic segmentation. We divide these approaches into a taxonomy of five distinct categories, listed in Table III – **encoder-decoder, multi-branch, meta-learning, attention and training pipeline**. While some works may straddle two or more of these categories, we assign them based on what we perceive to be the most significant contribution of the work. Works are presented in chronological order to provide an overview of how each paradigm has evolved.

### 4.1. Encoder-Decoder

Semantic segmentation requires that structural information be maintained between input and output, however many CNN architectures addressing this task have their origins in classification, targeting such datasets as Imagenet [48], where this is not the case. Discarding the final fully-connected layers from such an architecture leaves feature maps that retain some spatial information, however these tend to encode a very low-resolution representation due to the repeated downsampling operations required to increase receptive field while maintaining reasonable memory usage. Encoder-decoder architectures are a common approach to addressing this problem, whereby this low-resolution encoding is 'decoded' by an upsampling network that generates an output with the same dimensions as the input image. In most cases it is feature maps that are upsampled, with the class probability map only generated at the final output resolution, however one notable exception to this is FCN [17] in which the class probabilities are generated at low resolution and subsequently upsampled. This upsampling may be performed by interpolation operations interspersed with learned convolutions or by fractionally-strided convolutions that learn an optimized upsampling function.

A prominent early example of such an architecture applied to semantic segmentation of driving scenes is **SegNet** [18], depicted in Fig. 6, originally proposed in 2015. Built upon a relatively inefficient VGG16 [49] encoder and mirror-image decoder, real-time inference speed was not a priority, however compared to some of the bulky architectures that have since achieved state of the art results [50], SegNet can be considered relatively compact. One notable novelty of the work is the transferal of max-pooling indices – i.e. the location within its receptive window from which the pooling operation takes its output value – from encoder to decoder, enabling more accurate reconstruction of feature maps during upsampling operations, however this technique has not been widely utilized in subsequent works.

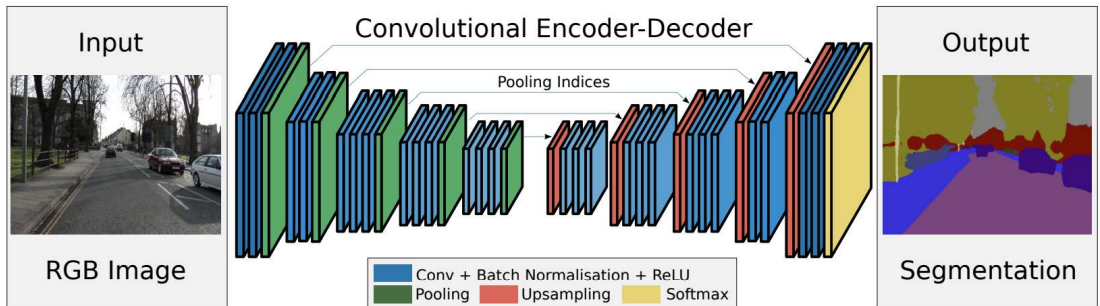

Fig. 6. Overview of the SegNet [18] encoder-decoder architecture for semantic segmentation.

**U-net** [1] is a compact encoder-decoder architecture that utilizes skip connections between corresponding encoder and decoder layers to retain high resolution features. These skip connections concatenate encoder and decoder feature maps at each scale, resulting in a relatively computationally expensive decoder due to the additional feature depth, however this technique combined with a novel boundary-weighted loss function achieved state of the art results in the medical image segmentation task for which u-net was designed.

One of the earliest prominent works specifically addressing real-time semantic segmentation was **Efficient Neural Network (ENet)** [34], proposed in 2016. Comprising a (relatively) large encoder and very simple decoder, ENet is built out of several variations of bottleneck residual blocks comprising a dimensionality reduction, convolution, and expansion. The convolution can be a standard 3×3, asymmetric 5×5, dilated, or fractionally strided in the case of the decoder. In the case of a downsampling block, the reduction operation is 2x2 convolution with a stride of 2, while a max-pooling operation is applied to the residual connection. Input is aggressively downsampled early in the network, which combined with the efficient use of convolutions, results in a very compact and fast architecture viable for deployment on embedded devices.

Treml et al [35] proposed an efficient segmentation architecture (**SQNet**) based on a Squeezenet [43] backbone, utilizing 'Fire modules', in which a 1×1 convolution reduces, or 'squeezes', feature maps to decrease the number of parameters required by subsequent parallel 1×1 and 3×3 convolutions whose outputs are concatenated. The bottleneck utilizes parallel convolutions with different dilation rates, based on the Deeplab architecture [50], to increase receptive field. Finally, an decoder comprising alternating fractionally strided convolutions and refinement modules, in which upsampled features are concatenated with those from corresponding decoder layers via skip connection, generates the final segmentation. Results demonstrate marginal improvement over ENet, however a slightly slower inference speed.

In 2017, Romera et al [51] introduced the **Efficient Residual Factorized Convnet (ERFNet)**, an encoder-decoder architecture comprising 1D factorized residual blocks, in which each 3×3 convolution is replaced by a 3×1 and 1×3 convolution, reducing the required parameters by one third. Each block contains two such pairs, built into an architecture similar to that of ENet, with several residual downsampling blocks deployed early in the network and a decoder that is much smaller than the encoder. Compared to ENet, ERFNet was demonstrated to produce significantly more accurate outputs at the cost of a slightly slower inference speed.

**Linknet** [52] utilizes a u-net-like architecture, with similarly sized encoder and decoder with skip connections at each scale, although encoder and decoder features are elementwise added, rather than concatenated as in u-net. Encoder blocks comprise four 3×3 convolutions, the first of which uses a stride of 2 to downsample its input, and a residual connection, while decoder blocks comprise an up-sampling 3×3 fractionally strided convolution in between 1×1 reduction and expansion convolutions. Linknet demonstrated significant improvement over prior real-time approaches on the Cityscapes dataset.

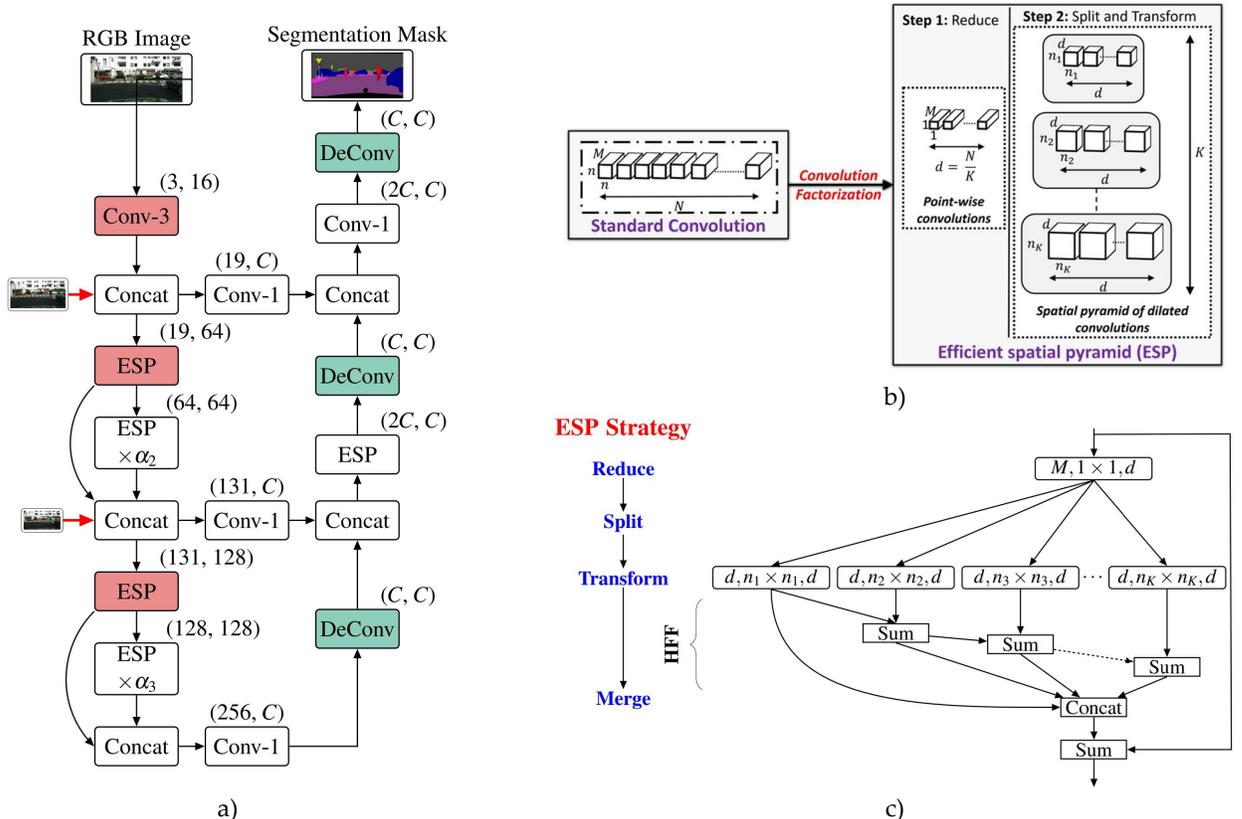

Fig. 7. a) Overview of the ESPNet [57] architecture. Red blocks downsample, green blocks upsample, numbers in brackets are input and output channels, where *C* is the number of output classes. b) illustrates how a standard convolution is decomposed into an ESP module. c) The operations that make up an ESP module: 1x1 convolution reduces dimensionality of the feature map that is then split between convolutions with different dilation rates. Outputs are merged by both sum and concatenation to reduce dilation artefacts, and then summed with the input.

In 2018, Siam et al presented a framework for real-time semantic segmentation (**RTSeg**) [53] facilitating the comparison of multiple different backbone encoder networks and meta-architectures. After comparing VGG16 [49], ResNet18 [42], MobileNet [45] and ShuffleNet [40] backbones, implemented in u-net [1], SkipNet (FCN8s) [17], and Dilation Frontend [54] architectures, the best combinations were found to be SkipNet-MobileNet and SkipNet-ShuffleNet, for overall accuracy and optimal compactness/accuracy respectively. Dilation architectures were shown to not perform as well, and while u-net performance was comparable to SkipNet, it resulted in much larger, slower networks.

Neskarov et al [55] proposed a **lightweight version of Refinenet (LWRF)**. The original Refinenet [56] refines the output of a backbone network at multiple scales using Refinenet Blocks, comprising simplified residual blocks, multi-resolution fusion, in which smaller feature maps are upscaled and summed with larger maps, and chained residual pooling blocks, in which features are extracted at successively smaller scales via pooling then upsampled and summed as an efficient method for increasing receptive field. The proposed lightweight version of Refinenet discards the simplified residual blocks and replaces the original 3×3 convolutions in both Fusion and Chained Residual Pooling blocks with kernels of 1×1. These changes facilitate a significant decrease in the number of parameters and inference time compared to the original Refinenet with only a small decrease in accuracy.

The **Efficient Spatial Pyramid** module of **ESPNet** [57], proposed by Mehta et al in 2018, uses a 1×1 convolution to reduce input dimensionality followed by parallel convolutions with different dilation rates to increase the receptive field. To avoid grid artefacts caused by the different rates of dilation, outputs are hierarchically summed, with the results concatenated and finally added to the input via residual connection. Fig. 7. outlines the design of the ESP module as well as the overall architecture, which utilizes ESP modules at multiple scales in the encoder as well as in a lightweight decoder. At each scale, the downsampled input image is concatenated with the encoder feature map and passed to both the next ESP module as well as to the corresponding decoder layer via skip connection. This results in a very compact network capable of deployment on embedded devices, with performance better than any comparably compact architecture at the time. In 2019, the authors proposed **ESPNetV2** [58], introducing the Extremely Efficient Spatial Pyramid (EESP) module. This increases the efficiency of the ESP module by using a grouped convolution for the initial dimensionality reduction, with each of the output grouped feature maps being passed to one of several depth-wise separable dilated convolutions. An additional 1×1 grouped convolution is applied after the outputs are concatenated. Due to the increased efficiency of the EESP module, a greater number of them can be used in the network, leading to a significant increase in performance for a comparable network size.

The authors of **Swiftnet** [59] argue that standard CNN architectures pretrained on large-scale classification datasets such as Imagenet [48] are well suited to being the backbone of real-time semantic segmentation networks. The output of this backbone is passed to a Spatial Pyramid Pooling block, a simplified version of the Pyramid Pooling Module of PSPNet [60], which aims to increase the network's receptive field. A very simple decoder then upsamples by interpolation, elementwise adds the upsampled features with the output of the corresponding encoder layer via skip connection, and applies a single 3×3 convolution at each scale.

**Fast Semantic Segmentation Network (Fast-SCNN)** [61] is a very lightweight architecture that is able to deliver good segmentation results at extremely high framerates. Input is aggressively downsampled via an initial three strided convolutions, with the resulting feature map passed to two branches. The low-resolution branch comprises several residual bottleneck blocks followed by a pyramid pooling module [60] to extract global context. The high-resolution branch comprises a single convolution block, with the outputs of both branches summed and passed through a very simple decoder comprising three convolution blocks and upsampling to compute the final output. Most of the convolutions are made depth-wise separable to increase efficiency.

**Depth-wise Asymmetric Bottleneck (DABNet)** [62] introduced the DAB module, which increases efficiency through the combination of depth-wise separable and asymmetric factorized convolutions. Each block begins with a standard 3×3 convolution, followed by two branches, each of which comprises depth-wise 3×1 and 1×3 convolutions, one of which is dilated. The outputs of these two branches are summed and passed to a final 1×1 convolution, whose output is summed with the block's input via residual connection. Minimal downsampling is used, as the receptive field is increased via dilation, and the input image is concatenated to the feature map at each scale. The class probability map is generated at the smallest scale of the encoder and upsampled by interpolation, and so DABNet may not truly be considered an encoder-decoder architecture, due to the lack of solution. However, in 2020 the authors proposed **Pointwise Aggregation Decoder (PAD)** [63], which adds a lightweight decoder to the DAB encoder. The PAD aggregates representations from different scales within the encoder, combining them to generate the final full-size representation from which the output is generated. The addition of a decoder provides a small boost to the accuracy of the output with minimal impact on inference time.

The authors of **ShelfNet** [64] propose a densely connected encoder-decoder-encoder-decoder architecture. The first encoder is a ResNet [42] backbone with features output at each scale and passed to the corresponding block of the first decoder, via a 1×1 convolution that reprojects output to the required number of channels. Both decoders consist of an identical structure of alternating fractionally strided convolutions to upsample between scales, and 'S-Blocks', which are modified residual blocks. Each S-Block first performs an elementwise add to combine the upscaled output of the previous S-Block with the output from the corresponding encoder block, followed by two 3×3 convolutions that share weights, approximating a simple recurrent network while halving the number of parameters compared to a standard residual block. During training, dropout is applied between these two convolutions to reduce overfitting. Finally, the input is added to the output of these convolutions via residual connection. The second encoder uses a similar structure of S-Blocks interspersed with strided convolutions to downsample. The output of each S-Block in the first decoder is passed to the corresponding S-Block of the second decoder, whose output is subsequently passed to the corresponding S-Block of the second decoder. This densely connected structure facilitates a large number of paths data can take from input to output, which can be considered an efficient ensemble of encoder-decoder architectures.

**Efficient Dense Asymmetric (EDAnet)** [65] proposes an EDA module that consists of a 1×1 convolution that reduces feature dimensionality followed by two asymmetric convolution pairs, the latter of which may be dilated. Module input is concatenated with the output features, and modules are densely connected within a larger block so that information can be shared across a wider receptive field. Input is downsampled by strided convolution and concatenation with a parallel max-pooling operation, however no decoder is used, with the class probability map computed at 1/8 scale and upsampled by bilinear interpolation, as relatively little downsampling is required due to the additional receptive field of the dilated convolutions used in later EDA modules.

**Efficient Ladder-Style DenseNets (LDN)** [66] employ a DenseNet [67] backbone within a Ladder network [68] architecture to achieve highly accurate results at reasonable inference speed. A DenseNet is built out of dense residual blocks in which the input to each convolution unit is the concatenated outputs of all prior units within the block as well as the block's input. This in contrast to a ResNet, where each convolution unit receives the elementwise sum of the prior unit's output and the block's input. A Ladder network is a u-net style encoder-decoder with skip connections where each decoder block upsamples the output of the prior block via interpolation, performs elementwise addition with the output of the corresponding encoder block and passes the ouput through a convolution unit before passing it to the next decoder block. In both encoder and decoder, a convolution unit comprises a 1×1 and a 3×3 convolution, with each encoder block containing four such units. To increase receptive field, a Spatial Pyramid Pooling block [60] is applied at the bottleneck, in which features extracted at four different scales are concatenated before being passed to the first decoder block. During training, auxiliary losses are applied at each scale of the decoder to aid learning.

The authors of **Real-Time General Purpose Semantic Segmentation (RGPNet)** [69] propose a densely connected encoder-decoder architecture based on a ResNet [42] backbone with a lightweight encoder, and a novel 'adapter' that sits between the two. Each adapter block takes the output of the corresponding encoder block, the adapter block at one scale larger and the decoder block at one scale smaller. Each input goes through a convolution, with stride applied as necessary to upsample or downsample accordingly, before an elementwise addition combines the outputs. A final residual block that shares weights between all scales then computes the feature maps that are passed to the decoder. This strategy for combining features at multiple scales results in excellent performance, and while real-time performance is demonstrated for the ResNet18 based RGPNet, the model is quite large and slow compared to some of the other approaches discussed.

### 4.2. Multi-Branch

The combination of global and local information is an important challenge in the pursuit of accurate semantic segmentation. Encoder-decoder based approaches usually address this by progressively downsampling feature maps, extracting increasingly high-level information at each scale. One major challenge in this approach is the preservation of high-resolution details extracted early in the network, and so several multi-branch architectures have been proposed that aim to address this by extracting features at different scales independently. For the purpose of this taxonomy, we consider a work to be a multi-branch model if the original input image is fed to the network at two or more scales.

In 2018, Zhao et al proposed **Image Cascade Network (ICNet)** [70], comprising three encoder branches, operating on ¼, ½ and full-sized input images. The first branch extracts semantic information using a deep PSPNet [60], with the low input resolution facilitating real-time inference from what is usually a relatively cumbersome architecture. The mid- and high- resolution branches utilize a simple lightweight architecture to extract more fine-grained information for refining output boundaries. Each branch downsamples its input by a factor of 8, with outputs combined by two 'Cascade Feature Fusion' (CFF) units. Each CFF takes input at two scales, with the lower resolution feature map upsampled by bilinear interpolation and a single dilated convolution before elementwise summation with the higher resolution feature map. Auxiliary losses are applied at each CFF unit, and the final output is upsampled by interpolation.

**ContextNet** [39] utilizes two branches to efficiently extract both spatial and contextual features. The contextual branch takes a ¼ resolution input image and is relatively deep, comprising twelve residual bottleneck blocks utilizing depth-wise seperable convolutions, sandwiched between two standard convolutions. The spatial branch comprises just one standard convolution and three depth-wise seperable convolutions. Output of the context branch is upsampled and passed through a dilated depth-wise convolution before outputs go through an elementwise addition, with one final convolution unit computing the class probability map. During training, the depths of all feature maps are doubled, with pruning [71] used to optimally reduce them to their original intended dimensions, and an auxiliary loss is applied to the output of the context branch to aid learning.

**Guided Upsampling Network (GUN)** [72] features two branches, taking input at ¼ and ½ scale, based on the Dilated ResNet [42] architecture. The higher resolution branch only uses the first few layers, which share their weights with those of the low-resolution branch, and the outputs are combined in a 'Fusion Module' comprising an elementwise summation and several convolutions that progressively reduce the number of feature maps. A second Fusion Module combines the result with the output of the first layer of the higher-resolution branch so that fine-grained details are retained. The main contribution of the work is the 'Guided Upsampling Module' (GUM), in which a separate branch learns to predict pixel offsets to guide the upsampling operation that generates the final output. Unlike nearest neighbor interpolation, where a pixel value is applied naively over a fixed kernel, the generated offsets are used to more accurately propagate pixel values over the upsampled output.

**Bilateral Segmentation Network (BiSeNet)** [19] is another two-branched architecture, consisting of a context path that takes input at ¼ of its original dimensions, and a spatial branch that takes full resolution input, illustrated in Fig. 8. The spatial path is very simple, utilizing just three strided convolutions along with batch normalization and rectified linear activation functions, while the context path is based on a more complex Xception architecture [38]. 'Attention Refinement Modules' (ARM), inspired by Parsenet [73] are applied at the outputs of the final two stages of the context branch, employing global average pooling to generate a feature vector the encodes global context, which is then reshaped such that it can be elementwise multiplied with the module input via residual connection. The outputs of the two ARMs are combined with the spatial branch output via concatenation within a 'Feature Fusion Module' (FFM), in which another residual global pooling block is used to encode global context. The output is upsampled to generate the final class probability map, with an auxiliary loss also applied to the output of the context branch. Overall, BiSeNet is an architecture that prioritizes speed, demonstrating results that are not quite as accurate as those of some of the slower discussed approaches.

In 2021, the authors of the original BiseNet paper proposed **BiSeNetV2** [74], demonstrating significant improvements in both speed and performance over the original. The context branch now combines the Stem Block of Inception-ResNet [75], using strided convolutions with pooling residual connections to aggressively downsample, with MobileNetv2 [46] Inverted Bottleneck blocks, which utilize depth-wise separable convolutions to increase efficiency. The FFM is replaced by a 'Bilateraly Guided Aggregation Layer', in which the output of each branch is sent along two branches, comprising either depth-wise separable or standard convolutions, with the outputs combined by a combination of elementwise product and elementwise sum. During training, auxiliary loss is applied at several scales along the context branch, and Online Hard Example Mining [76] is used to improve training on difficult samples.

### 4.3. Meta-Learning

Here we use the term 'Meta-Learning' to refer broadly to techniques where a learned function directly affects the architecture being used for the task at hand. Most examples of meta-learning in the field of real-time semantic segmentation come under the category of Neural Architecture Search (NAS) [77], which is a method for automating the process of designing a neural network architecture. Typically, it involves determining a search space, a search algorithm, and an optimization function. NAS often just involves finding an architecture that delivers optimum results, however in a real-time setting, architecture size, complexity, and inference time form additional factors that should be considered in the optimization function. We also include the concept of the Hypernetwork [78], in which a network learns to set the weights of another network, in this category.

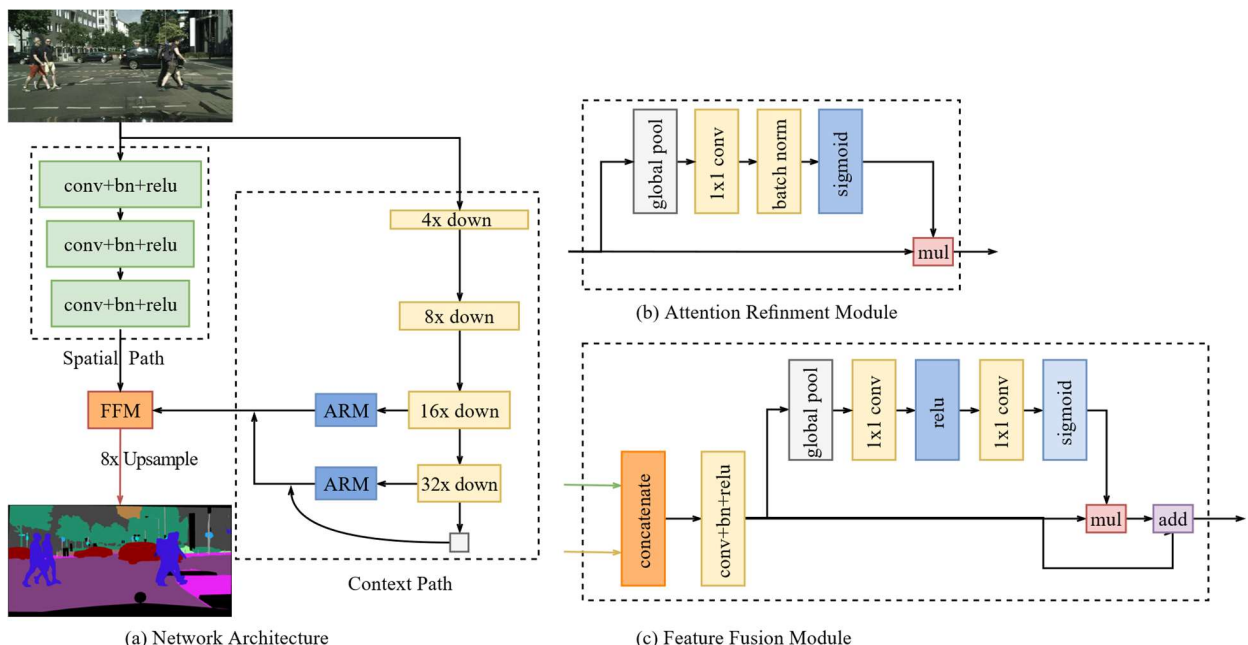

Fig. 8. a) Overview of the BiSeNet [19] architecture, comprising spatial and context paths. b) Attention Refinement Module (ARM) that extracts global information at the end of the context path. c) Feature Fusion Module (FFM) that combines spatial and context information to compute the final output.

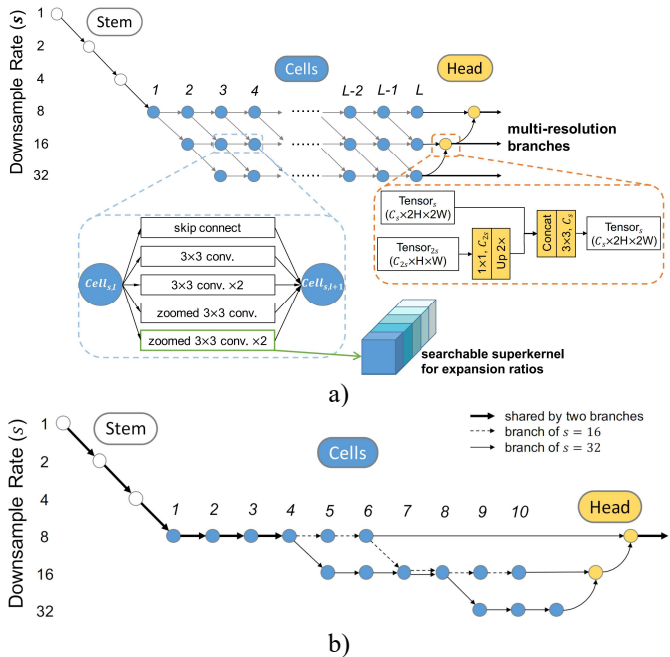

Fig. 9. a) The Fasterseg [83] supernetwork, representing all possible networks that could be chosen by the architecture search algorithm. b) The architecture that was selected by the algorithm when optimising for speed and performance on the Cityscapes [28] dataset.

In 2019 **SqueezeNAS** [79] proposed a supernetwork-based approach to learning an architecture for real-time semantic segmentation, whereby an architecture is built of a fixed number of inverted residual blocks, each of which has a fixed input size, output size and stride, and can comprise one of thirteen configurations. These configurations vary channel grouping, convolution dilation, kernel size and expansion ratio, with a residual connection as the thirteenth possible configuration, and a Gumbel-Softmax classifier [80] is used at each block to select which configuration should be used while maintaining differentiability. A lookup table that lists the latency of each possible configuration for each block on a target hardware platform, in this case an Nvidia Xavier embedded system [81], is used in addition to segmentation accuracy to ensure the final architecture is efficient. An ASPP [60] module, or LR-ASPP [82] in a more efficient version of the proposed architecture, is utilized as decoder to create the final output. The training process involves first training the supernetwork to select both architecture and weights for the target task, then once the architecture is selected the weights are randomized so that the chosen network can be trained from scratch, including any pretraining.

**FasterSeg** [83], illustrated in Fig. 9., utilizes a very large search space wherein a multi-resolution branching architecture must be selected as well as the configuration of each block comprising each branch. During search, three parameters encode each block's selection of operator, input and expansion ratio respectively. The operator can be either one or two standard convolutions, one or two "zoomed convolutions", in which the convolution is sandwiched between a downsampling and an upsampling operation to increase receptive field while reducing parameters, or a skip connection, in which case the block is discarded. A probability is assigned to each potential operator and the greatest is selected. A block's input may be the output of the prior block on the same branch or the downsampled output of the prior block on the prior branch. A probability is assigned for each, and for each branch the single block with the highest probability of taking input from the prior branch is selected to do so, while prior blocks are discarded. Expansion ratio is selected from a set of predefined values, with Gumbel-Softmax [80] applied to enable a discrete selection while maintaining differentiability. Optimization of the searched architecture considers latency via a fine-grained lookup table, with the different dimensions of the search space considered independently. The searched branches are preceded by a sequence of strided convolutions to extract low level features and downsample the input, while a sequence of upsampling, concatenation and convolution combines branch outputs and generates the final segmentation. Distillation is integrated into NAS via a novel approach of simultaneously searching for teacher and student networks, whereby the teacher network is derived from the same search space but without the same latency constraints.

**Graph-Guided Architecture Search (GAS)** [84], published in 2020, propose a semantic segmentation architecture comprising a sequence of cells that are configured by graph neural networks [85]. Each cell can select one of eight possible operations: depth-wise separable 3×3 convolution with dilation 1, 2, 4, or 8, standard 3×3 convolution, max pooling, skip connection or zero operation. The selection is made via a directed acyclic graph in which each edge represents a single candidate operation as a one-hot vector, and Gumbel-Softmax [80] is used to make the discrete selection differentiable. Adjacent network cells communicate with each other via a 'Graph Convolution Network (GCN)-Guided Module' (GGM), in which matrices encoding the probability of each operation at each edge of the current and previous cells are passed to a GCN [86] that learns to model information propagation between the two cells. The architecture is optimized by a weighted combination of cross-entropy loss between output and ground truth and the latency of the selected operations on a target GPU, taken from a lookup table, with the final selected architecture subsequently trained independently without the latency loss. The overall architecture consists of strided convolutions that downsample input, the sequence of graph-searched cells, and an ASPP [60] module that computes the output.

In 2021, **Hyperseg** [87], illustrated in Fig. 10., proposed an efficient semantic segmentation architecture that combines a hypernetwork with locally connected patch-wise convolutions. An Efficientnet [47] backbone encoder extracts features, which are passed to a hypernetwork decoder via u-net [1] style skip connections that apply a 1×1 convolution to reduce channels for efficiency. The final output of the backbone is passed to a 'Context Head', which uses a small nested u-net to generate a signal that is used by weight mapping blocks in the decoder to generate the weights used in its convolutions. The decoder is made up of several 'Meta Blocks', based on the inverted residual blocks of MobileNetV2 [46], however the weights of the convolutions within these blocks are determined at runtime by a corresponding weight-mapping block. These dynamic weights are also patch-wise, meaning different sets of weights are applied to different image regions. Grouped convolutions are used to reduce the number of weights that need to be computed.

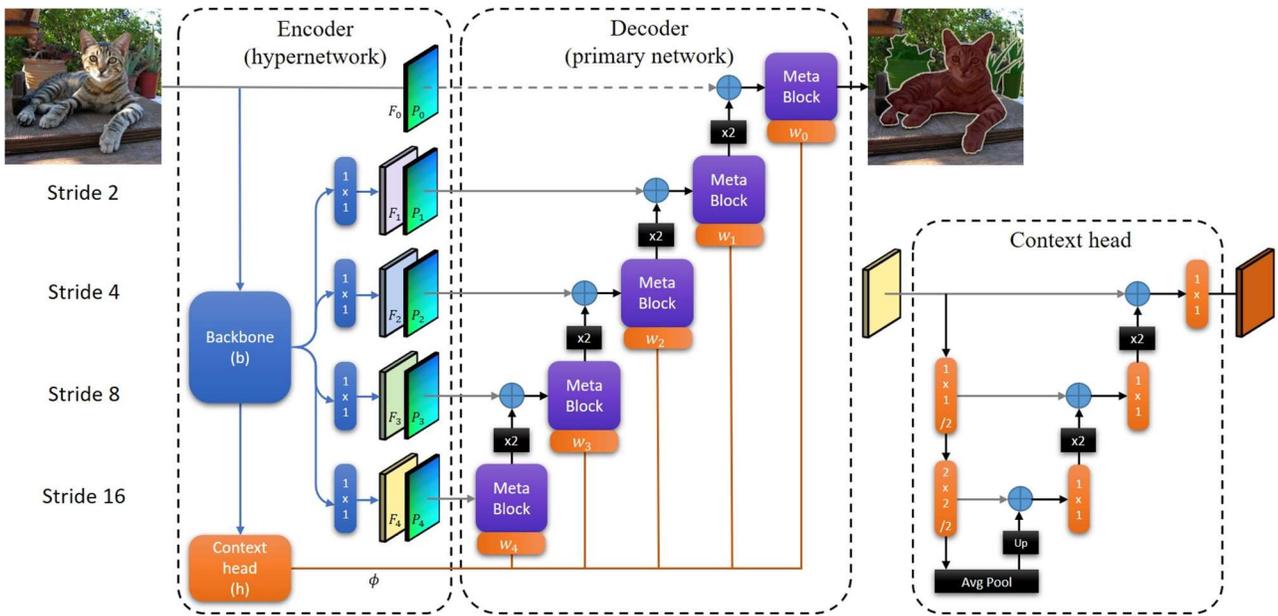

Fig. 10. Overview of the architecture of Hyperseg [87]. The encoder passes feature maps at multiple scales to corresponding decoder blocks, the weights of which are generated during inference by weight generator modules whose input comes from a Context Head placed at the end of the encoder.

### 4.4 Attention

Attention-based models, originally proposed for language-based tasks [88], learn to prioritise the most relevant parts of an input sequence. Self-Attention in particular has been shown to be a useful technique applied to vision-based tasks [89], however it can lead to highly cumbersome and inefficient models due to the quadratic complexity of computing the relationship between every pair of data points, which in vision models usually means pixels, which can number in the millions. Attempts to reduce this complexity include patch-, rather than pixel-, wise attention [90], and axial attention [91], in which relationships are separately computed across image rows and columns, although these are still not immediately suited to real-time inference due to computationally expensive softmax functions that number in the thousands.

**Deep Feature Aggregation (DFANet)** [36], proposed in 2019, is an encoder-decoder architecture with 'FC Attention' modules placed between the encoder and decoder. The encoder comprises three lightweight Xception [38] branches, with each block output concatenated with the input to the corresponding block of the next branch, while the decoder is a relatively simple sequence of convolution and upsampling operations that fuses features from multiple encoder scales. When a classification architecture is adapted for a segmentation task, the final fully-connected (FC) layers, which are able to aggregate context across whole images at the expense of structural information, are typically discarded. DFANet, however, includes an FC

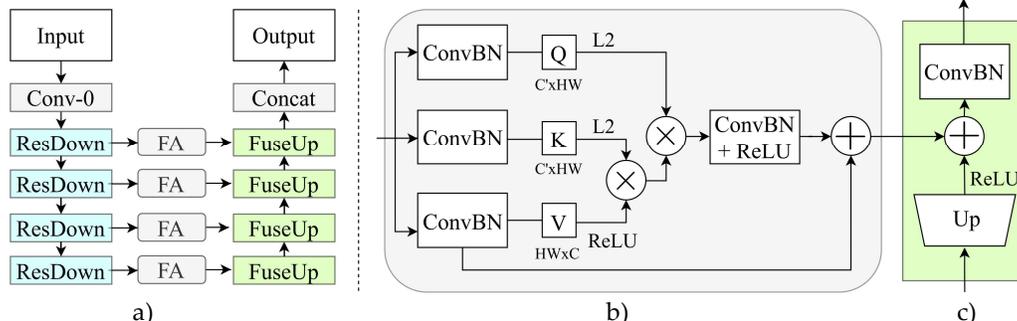

Fig. 11. a) Overview of the Fast Attention [93] network architecture, with Fast Attention (FA) blocks utilized in the skip connections between encoder and decoder. b) FA block, in which pointwise convolutions generate the Query, Key, and Value matrices from which attention is computed. c) FuseUp module of the decoder that combines FA outputs at different scales.

Attention module at the end of each of its encoder branches, which exploits the FC layer's capability to encode relationships between all pixels in a feature map, while structural information is retained via residual connection wherein the FC output is reprojected to match the input shape, and the two are combined by elementwise multiplication and passed both to the first block of the next encoder branch, and to the corresponding decoder block. The first half of the decoder fuses the outputs of early encoder blocks by elementwise sum, while the latter half fuses these with the FC Attention outputs.

**Lightweight Encoder-Decoder (LEDNet)** [92] proposes an 'Attention Pyramid Network' as a decoder in which a single output is taken from the encoder and passed to multiple decoder branches at different scales, including a global pooling branch, which uses global average pooling to generate feature maps of dimensions 1×1 that capture global context. Branch outputs are then combined by either pointwise sum or pointwise product and upsampled to compute the final segmentation. A ResNet [42] backbone is utilized for the encoder, with standard residual blocks replaced by 'Split-Shuffle-non-Bottleneck'(SS-nbt) blocks: Input channels are split into two groups, each of which is passed to a separate branch comprising asymmetric and dilated convolutions; The two branch outputs are concatenated, then the elementwise sum is computed with the original block input via residual connection, and finally the channels are shuffled to facilitate the sharing of information between branches.

**Fast Attention (FAnet)** [93], from 2020, proposes an efficient version of Self-Attention that extracts global context in the skip connections between encoder and decoder, as shown in Fig. 11. As in standard Self-Attention, pointwise convolutions are applied to the flattened feature map to compute Value, Query and Key matrices, however L2 normalisation is used in place of softmax in the affinity operation which is itself more efficient, but which also allows the order of matrix multiplications to be changed for greater efficiency. A single convolution reshapes the output so that it can summed with the Value matrix. These FA blocks are employed in an encoder-decoder architecture with a ResNet18 [42] backbone with additional downsampling for increased efficiency, with each residual block passing its output to a corresponding FA block which passes its output to the corresponding decoder block, where it is concatenated with the output from the previous decoder block.

The **Bilateral Attention Decoder (BAD)** [94], published in 2021, applies two different attention mechanisms to high- and low- resolution feature maps, respectively spatial attention and channel attention. Spatial attention uses a pointwise convolution to reduce channels in the feature map to 1, followed by parallel average and max pooling operations, each with a stride of 1 (i.e. the output size does not change), the outputs of which are concatenated and again pointwise convolved back down to 1 channel. The final output is a feature map of the same dimensions as the input, computed as the pointwise product of the single-channel feature map and each channel of the input feature map. Channel attention uses global average pooling to generate a feature map comprising a single value for each input channel, which then goes through a pointwise convolution before being multiplied with the input feature map. The high- and low-resolution attention features are fused in a 'Pooling Fusion Block', in which average pooling is used to smooth the inferred class boundaries within the upsampled low-resolution feature map before concatenation with the high-resolution feature map which should contain the fine-grained detail required to infer the true boundary. The overall architecture used comprises a ResNet18 [42] encoder and a decoder made up of two BAD blocks at different scales, with high-resolution features passed via skip connection from the corresponding encoder blocks. During training, auxiliary loss is used on the BAD output in addition to the main loss function applied to the final network output.

### 4.5 Training Pipeline

The final category of works we discuss are those that take an existing architecture and change the training process so that more accurate results can be achieved. One widely used technique for improving the performance of compact models without changing their architecture is knowledge distillation [95], whereby the outputs of a large, high-performing "teacher" model are used to train the smaller, efficient "student" model. By more closely matching the output distribution of a more capable model, the student is able to capture nuances within its training data that it would fail to learn from ground truth labels alone.

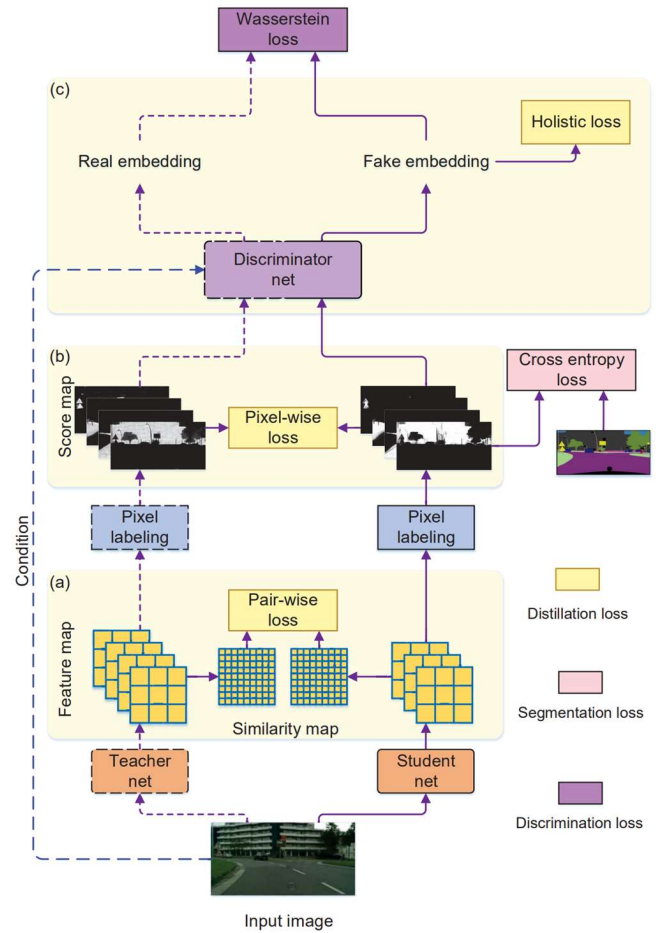

Fig. 12. Overview of the Structured Knowledge Distillation [96] training pipeline, combining cross entropy, pair-wise, pixel-wise and holistic loss functions.

**Structured Knowledge Distillation** [96], proposed in 2019, builds on the idea of distillation in the context of semantic segmentation via pixelwise, pairwise and holistic comparison between a compact student and a larger PSPnet101 [60] teacher model. The process for training the student model is illustrated in Fig. 12. Pixelwise distillation seeks to teach the student model to better match its output class probability maps to those of the teacher by computing the KL Divergence between pixel values.; Pair wise optimization computes the similarity between each pair of pixels in a feature map, with the resulting similarities compared between teacher and student; Holistic distillation trains a separate discriminator network, based on Self Attention GAN [97], to differentiate between the output class probability maps of teacher and student networks, with the aim that the student minimizes the Wasserstein distance [98] between the resulting embeddings. Student output is also compared to the ground truth segmentation via cross-entropy loss. Significant improvement in performance is demonstrated for several efficient student architectures with no impact on inference runtime.

**Knowledge Adaptation** [99] proposes two new types of distillation that exploit the structural information inherent in semantic segmentation, "Knowledge Adaptation" and "Affinity Distillation". Knowledge Adaptation passes the final feature maps from the teacher to a separate autoencoder that aims to learn to recreate its input via a compact latent representation. A "Feature Adapter" takes the final feature maps of the student model and aims to produce an output that matches this latent representation, such that the student better learns to approximate the feature encoding of the teacher. Affinity Distillation computes the affinity between every pair of pixels in the final feature map as the matrix product of the feature vectors of each pixel pair, with the aim of minimizing the squared distance between the resulting affinity matrices of teacher and student. The teacher model combines a ResNet-50 [42] backbone with ASPP [60], while several variations of MobileNetV2 [46] are used as student, demonstrating a significant increase in performance with no effect on inference latency.

In 2021, Fan et al [100] proposed the **Short-Term Dense**

TABLE 4

COMPARISON OF REAL-TIME SEMANTIC SEGMENTATION TECHNIQUES

| Publication Year | Model | Parameters (million) | Cityscapes | | CamVid | fps - RTX 3090 | | | fps - Jetson |
|---|---|---|---|---|---|---|---|---|---|
| | | | Val | Test | Test | 2048×1024 | 1024×512 | 512×256 | 512×256 |
| 2015 | SegNet [18]* | 1.4 | - | 56.1 | - | 9.2 | 34.4 | 126.5 | 6.4 |
| 2016 | ENet [34]* | **0.37** | - | 58.3 | 51.3 | 45.5 | 107.7 | 142 | 28 |
| | SQNet [35]* | - | - | 59.8 | - | 15.9 | 59.4 | 205.5 | 10.7 |
| 2017 | ERFNet [51] | - | - | 69.7 | - | 37.4 | 135.5 | 232.2 | 19.8 |
| | LinkNet [52] | 11.5 | 76.4 | - | 68.3 | 64.4 | 202.1 | 352.9 | 43.5 |
| 2018 | BiSeNet [19]* | 49 | 74.8 | 74.7 | 68.7 | 54.1 | 189.8 | 382.7 | 44 |
| | ContextNet [39]* | 0.85 | 65.9 | - | - | 177.3 | 319.7 | 336.1 | 77.1 |
| | ESPNet [57] | 0.4 | - | 60.3 | - | 88 | 330.7 | **412.9** | 74.6 |
| | ICNet [70]* | - | - | 70.6 | 67.1 | 111.3† | - | - | 6.1† |
| 2019 | DABNet [62] | 0.76 | 70.1 | 69.1 | 66.4 | 59.1 | 225.4 | 272.7 | 52.7 |
| | DFANet [36]* | 7.8 | 59.2 | 71.3 | 64.7 | 78.3 | 82 | 84.9 | 20.6 |
| | EDANet [65]* | 0.68 | - | 67.3 | 66.4 | 56.2 | 194.9 | 202.1 | 36.6 |
| | ESPNetV2 [58] | 0.79 | 66.4 | 66.2 | - | 32.5 | 114.7 | 140.2 | 32.4 |
| | FasterSeg [83] | 4.4 | 73.1 | 71.5 | 71.1 | 174.3 | 293.3 | 294.1 | **78.8** |
| | Fast-SCNN [61]* | 1.1 | 69.2 | 68 | - | **196.1** | **348.2** | 355.1 | 78.5 |
| | LEDNet [92] | 0.94 | - | 70.6 | - | 32.6 | 103.5 | 105.8 | 27.7 |
| | ShelfNet [64] | - | - | 74.8 | - | 69.5 | 146.3 | 181.7 | 23.3 |
| | SqueezeNAS-XL [79] | 3 | 75.2 | - | - | 48.3 | 168.6 | 239.1 | 40.2 |
| | SwiftNet [59] | 11.8 | 75.4 | 75.5 | 72.6 | 74.6 | 247.8 | 364.5 | 48 |
| 2020 | FANet-18 [93] | - | 75 | 74.4 | 69 | 147.7 | 308.7 | 313.5 | 63.6 |
| | FANet-34 [93] | - | 76.3 | 75.5 | 70.1 | 114.7 | 233.7 | 235.7 | 26.2 |
| | LDN [66] | 9.5 | **79** | **79.3** | - | 28.9 | 87.7 | 89.6 | 18.4 |
| 2021 | BiSeNetV2 [74]* | - | 75.8 | 75.3 | **78.5** | 68.1 | 239.3 | 273.7 | 31 |
| | Hyperseg-S [87] | 10.2 | 78.2 | 78.1 | - | 19.1 | 63.7 | 90.7 | 18.3 |
| | STDC1 [100] | - | 74.5 | 75.3 | - | 49.1 | 136 | 195.3 | 35 |
| | STDC2 [100] | - | 77 | 76.8 | 73.9 | 42.1 | 103.9 | 135.1 | 25.9 |

*evaluation was not conducted using authors' implementation   **Bold** indicates top result, red indicates < 30fps   †input size 2049×1025

Taxonomy: Encoder-Decoder   Multi-Branch   Attention   Meta-Learning   Train Pipeline

**Concatenate network (STDC).** Based on BiSeNet [19], STDC replaces the high-resolution spatial path with a 'Detail Guidance' module that is only utilized at train time to teach the early layers of the more efficient context path to extract the required spatial information. During training, features extracted early in the network are passed to a 'Detail Head' which aims to predict fine-grained boundary information, with ground-truth computed as the result of Laplacian kernels applied to the ground-truth segmentation at several different scales. The convolution blocks of BiSeNet are replaced by STDC modules, in which features are extracted at progressively lower dimensionality and concatenated to form a high-dimensional output feature map. These changes lead to a model with comparable performance to the original BiSeNet with significantly lower latency.

# 5 EVALUATION

In this section we present a quantitative comparison of prominent works addressing real-time semantic segmentation. For each approach evaluated, we quote, where available, the parameter count and performance on the Cityscapes [28] and CamVid [26] datasets from the original paper, specifically the mean intersection over union (mIoU) across the Cityscapes validation and test sets, and the CamVid test set. We also evaluate the runtime of each model at different resolutions under consistent conditions, using the authors' implementation where possible.

We evaluate inference speed under two scenarios: a high-end GPU research workstation, with an AMD Threadripper 3975 32 core CPU [101], 512GB RAM and an Nvidia RTX 3090 GPU [102], running Python 3.9 [103], Pytorch 1.9 [104], and CUDA 11.4 [105], on which we measure inference time at the full cityscapes resolution of 2048×1024 as well as half and quarter resolutions; and an Nvidia Jetson Xavier AGX Developer Kit [81], an embedded GPU system similar to the hardware that might be aboard an autonomous vehicle, running Python 3.6, Pytorch 1.8, and CUDA 10.2, on which we measure inference time at quarter resolution (512×256). We utilize the timing method of [59] whereby only model runtime is recorded, regardless of loading data in and out of GPU memory, and take the mean inference time over 100 forward passes of a randomly generated input tensor. Some of the evaluated works originally used the high-performance TensorRT framework [106] or 16-bit precision to boost inference speed, however we disregard these details to ensure a consistent comparison.

Table IV displays the results of our quantitative analysis, with works ordered by year of publication. We can see a steady increase in performance on both datasets over time, with LDN and BiseNet V2 demonstrating the best results on Cityscapes and CamVid test sets with 79.3 and 78.5 respectively. The best Cityscapes test result from a model capable of real-time inference at full resolution is 76.8 from STDC2. For context, the top result on the Cityscapes Leaderboard [107] for semantic segmentation not constrained by latency or memory limitations is 86.2, as of December 2021. Fast-SCNN demonstrates the fastest workstation inference time at full and half resolution, however does not appear to benefit much from further reducing input resolution, with the fastest inference time at quarter resolution coming from ESPNet. All works achieve real-time inference (>30fps) at half resolution, and all apart from SegNet, SQ, Ladder Dense Net and Hyperseg achieve the same at full resolution. The ICNet model is designed for a specific input size so we are unable to evaluate it at lower resolutions.

On the embedded GPU, Fasterseg attains the lowest latency, with almost half of the evaluated works not reaching real-time performance. The best Cityscapes test result from a model capable of real-time inference on our embedded system is 75.5 from swiftnet. In general, a work that is faster than another in the workstation scenario will be the same in the embedded scenario, however some works appear to handle the transition better than others, possibly due to specific hardware optimisations for the types of computations within a model.

Fig. 13 plots mIoU on the Cityscapes test set (validation set if not available) against workstation fps at full resolution of each evaluated work, with each point coloured according to the work's category in our taxonomy. We can see that, apart from a few earlier works, there is an inverse correlation between performance and speed, and that meta-learning approaches in particular appear to achieve a good tradeoff.

# 6 CONCLUSIONS

In this survey we have discussed many of the prominent works proposed to address the problem of low-latency semantic segmentation on memory-constrained hardware. We have discussed and categorized these works based on their major contributions to the field, namely encoder-decoder architectures, multi-branch networks, meta-learning, attention mechanisms and training pipelines. Finally, we have performed our own experiments to measure the inference speed of the discussed works under a consistent hardware and software environment, being, we believe, the first to do so for such a wide variety of models. We have shown that while there is generally a tradeoff between accuracy and speed, there are several works that achieve near state-of-the-art results with relatively compact models capable of real-time inference. We have also evaluated works on an embedded GPU system, demonstrating that real-time inference is possible on such a device, but that a model that is fast on a high-end workstation is not necessarily the same under constrained resources.

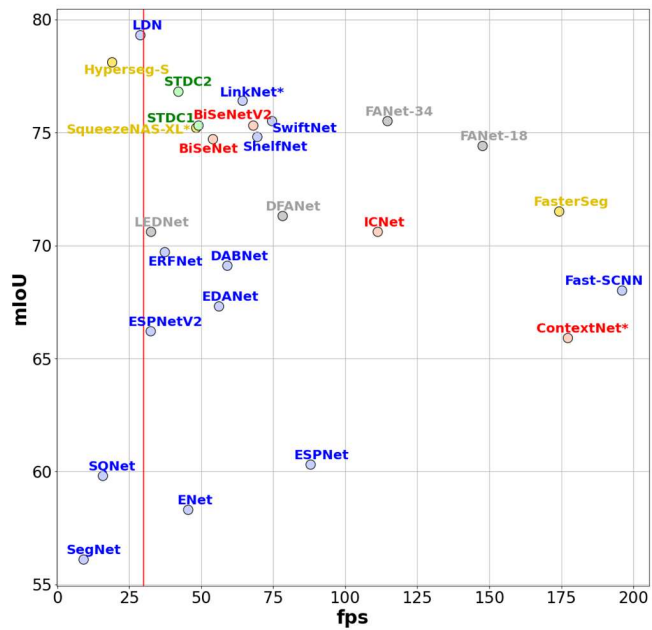

Fig. 13. Plot of performance against latency for each of the evaluated works. mIoU is computed on the Cityscapes [28] test dataset (* denotes validation set was used) and fps is computed on a GPU workstation at an input resolution of 2048×1024. Colour denotes a work's place within our taxonomy, as in Table 1. The red line marks 30fps, the threshold we use to consider inference to be in real-time.

One constant challenge in semantic segmentation is the fusion of global context information with fine-grained spatial detail, especially in the pursuit of efficient real-time models where low-resolution representations are favoured. The loss of high-resolution detail often leads to poorly defined class boundaries, with techniques proposed to address this, such as conditional random field [108] and multi-scale attention [109], often not feasible for low-latency, low-memory deployment. This remains a major challenge in the field, and something we believe future works will need to address if they are to surpass the current state of the art in real-time semantic segmentation.

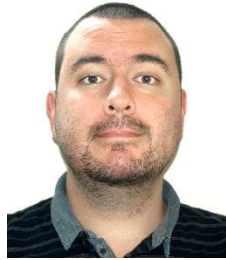

**Christopher J. Holder** received a BSc degree in Computer & Video Games from The University of Salford, UK, MSc degree in Computing from Cardiff University, UK, and PhD degree in Computer Science from Durham University, UK, focusing on the application of deep learning techniques to off-road autonomous driving. He has been a researcher at the Institute for Infocomm Research, Singapore, a postdoctoral researcher at Durham University, UK, Khalifa University, UAE and New York University Abu Dhabi, UAE, and is founder of Udara Limited, UK, an aerial imaging startup. His research focuses on the application of deep learning to visual problems, particularly localization, mapping, and scene understanding for robots and autonomous vehicles.

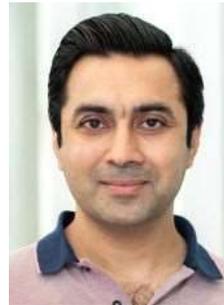

**Muhammad Shafique** (M'11 - SM'16) received the Ph.D. degree in computer science from the Karlsruhe Institute of Technology (KIT), Germany, in 2011. Afterwards, he established and led a highly recognized research group at KIT for several years as well as conducted impactful R&D activities in Pakistan and across the globe. In Oct.2016, he joined the Institute of Computer Engineering at the Faculty of Informatics, Technische Universitat Wien (TU Wien), ¨ Vienna, Austria as a Full Professor of Computer Architecture and Robust, Energy-Efficient Technologies. Since Sep.2020, he is with the Division of Engineering, New York University Abu Dhabi (NYU-AD), United Arab Emirates, and is a Global Network faculty at the NYU Tandon School of Engineering (NYU-NY), USA. His research interests are in brain-inspired computing, AI & machine learning hardware and system-level design, energy-efficient systems, robust computing, hardware security, emerging technologies, ML for EDA, FPGAs, MPSoCs, and embedded systems. His research has a special focus on cross-layer analysis, modeling, design, and optimization of computing and memory systems. The researched technologies and tools are deployed in application use cases from Internet-of-Things (IoT), smart Cyber-Physical Systems (CPS), and ICT for Development (ICT4D) domains. Dr. Shafique has given several Keynotes, Invited Talks, and Tutorials, as well as organized many special sessions at premier venues. He has served as the PC Chair, General Chair, Track Chair, and PC member for several prestigious IEEE and ACM conferences. Dr. Shafique holds one U.S. patent and has (co-)authored 6 Books, 10+ Book Chapters, 300+ papers in premier journals and conferences, and 50+ archive articles. He received the 2015 ACM/SIGDA Outstanding New Faculty Award, AI 2000 Chip Technology Most Influential Scholar Award in 2020, six gold medals, and several best paper awards and nominations at prestigious conferences like DAC, ICCAD, DATE, and CODES+ISSS.